\documentclass[10pt,twocolumn,letterpaper]{article}

\usepackage[pagenumbers]{iccv}              

\definecolor{iccvblue}{rgb}{0.21,0.49,0.74}
\usepackage[pagebackref,breaklinks,colorlinks,allcolors=iccvblue]{hyperref}
\usepackage{xcolor}
\usepackage{xparse}
\usepackage{amsmath}
\usepackage{amssymb}
\usepackage{physics}
\usepackage{bm}
\usepackage{overpic}
\usepackage{transparent}
\usepackage{customCommands}
\usepackage{multirow}
\usepackage{colortbl}

\def\paperID{14085} 
\def\confName{ICCV}
\def\confYear{2025}

\title{PhysPose: Refining 6D Object Poses with Physical Constraints}

\author{Martin Malenickýn$^{1}$\and Martin Cífka$^{1,2}$\and Médéric Fourmy$^{1}$ \and Louis Montaut$^{3}$ \and Justin Carpentier$^{3}$ \and Josef Sivic$^{1}$ \and Vladimir Petrik$^{1}$ \and \\
    \centerline{$^{1}$Czech Institute of Informatics, Robotics and Cybernetics, Czech Technical University in Prague}\\
    \centerline{$^{2}$ Faculty of Electrical Engineering, Czech Technical University in Prague}\\
    \centerline{$^{3}$ Inria - Département d’Informatique de l’École normale supérieure, PSL Research University}
}

\definecolor{tab:blue}{RGB}{31,119,180}

\newcommand{\mypar}[1]{\vspace{1mm}\noindent\textbf{#1}}
\newcommand{\myparcontinuous}[1]{\mypar{#1}\hspace{-0mm}}

\ExplSyntaxOn
\NewDocumentCommand{\sepcomma}{m}
{
  \seq_set_split:Nnn \l_tmpa_seq {} {#1}
  \seq_use:Nn \l_tmpa_seq {,}
}
\ExplSyntaxOff
\newcommand{\meas}[1]{\widetilde{#1}}

\newcommand{\itr}{} 

\NewDocumentCommand\sub{mg}{%
    _{#1%
    \IfNoValueF{#2}{,#2}%
    }
}

\makeatletter
\DeclareRobustCommand\onedot{\futurelet\@let@token\@onedot}
\def\@onedot{\ifx\@let@token.\else.\null\fi\xspace}
\def\eg{\emph{e.g}\onedot} 
\def\ie{\emph{i.e}\onedot}

\makeatother

\begin{document}
\maketitle
\begin{abstract}
  Accurate 6D object pose estimation from images is a key problem in object-centric scene understanding, enabling applications in robotics, augmented reality, and scene reconstruction. Despite recent advances, existing methods often produce physically inconsistent pose estimates, hindering their deployment in real-world scenarios. 
We introduce PhysPose, a novel approach that integrates physical reasoning into pose estimation through a postprocessing optimization enforcing non-penetration and gravitational constraints. 
By leveraging scene geometry, 
PhysPose refines pose estimates to ensure physical plausibility. Our approach achieves state-of-the-art accuracy on the YCB-Video dataset from the BOP benchmark and improves over the state-of-the-art pose estimation methods on the HOPE-Video dataset. Furthermore, we demonstrate its impact in robotics by significantly improving success rates in a challenging pick-and-place task, highlighting the importance of physical consistency in real-world applications.
        
\end{abstract}
\section{Introduction}

\begin{figure}[t]
    \centering
    \includegraphics[width=\linewidth]{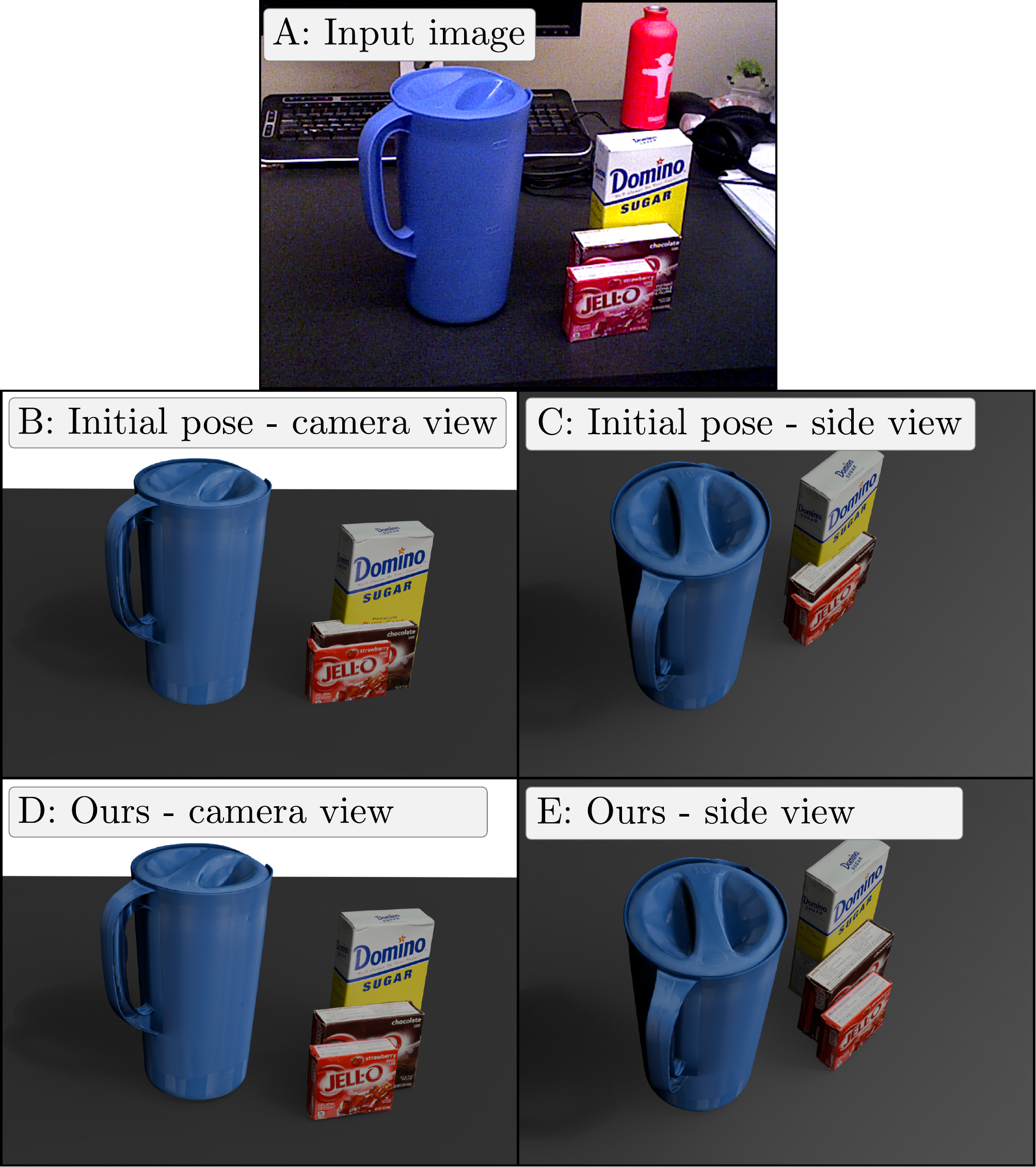}
    \caption{
    \textbf{Object pose estimation with physical consistency.}
    In an input image A (top), taken from the YCB-V dataset~\cite{xiang2017posecnn}, the state-of-the-art pose estimators (here~\cite{megapose}) predict 6D object poses that result in geometric penetrations, as shown in images B (camera viewpoint) and C (side view). Our PhysPose method resolves these physical inconsistencies, mitigating object interpenetration, as shown in images D and E.    
    }
    \label{fig:motivation}
\end{figure}

Accurately estimating the position and orientation (\ie, the 6D pose) of objects is a key research problem in embodied perception, with applications in autonomous driving, robotic manipulation, and augmented reality. 
Current state-of-the-art solutions for model-based object pose estimation~\cite{megapose, cosypose, foundationpose, ornek2024foundpose} do not explicitly take into account the geometric and physical constraints of the world.
This often results in geometric and physical inconsistencies, as illustrated in Fig.~\ref{fig:motivation}.
Consequently, objects may appear to float above surfaces where they should lie or may collide with other objects or furniture in the environment. 
These inconsistencies suggest the object's pose is inaccurately estimated, often causing task failures, such as the object not being grasped or being dropped during execution.

To overcome this challenge, we introduce a novel approach, which we call PhysPose, that integrates physical reasoning into pose estimation through a post-processing optimization.
Our technique enhances physical consistency within a scene by refining the per-object pose estimates obtained from existing methods.
Using readily available environmental geometry, often available in robotics for tasks such as motion planning, our framework further strengthens the physical consistency of the estimated scene.
For scenarios where scene geometry data might be limited, we have developed an automated onboarding process to estimate the support table's pose using only two views of the scene.
Given the static environment, this table pose estimation is a single, quick initialization step, easily performed by repositioning the robot's camera before the main operation without the need for any explicit calibration.
See qualitative results on our project page: \url{https://data.ciirc.cvut.cz/public/projects/2025PhysPose}.

\mypar{Contributions.} The key contributions of our approach are:
(i) A novel and general post-processing framework for enforcing physical consistency in object pose estimation, which incorporates scene geometry to refine pose estimates.
(ii) Benchmarking on two datasets, YCB-Video~\cite{he2017maskrcnn} and HOPE-Video~\cite{hopevideo}, demonstrating a significant consistent performance improvement over state-of-the-art single-view pose estimation methods.
(iii) Validation in a real-world RGB-based robotic pick-and-place task showcasing a notable improvement in task success rate, particularly when grasping near object edges.

\section{Related work}

\mypar{Image-based object pose estimation.} Estimating the pose of rigid objects from images is one of the fundamental problems in computer vision~\cite{roberts1961}.
Earlier methods relied on 2D-3D correspondences computed by specific-purpose algorithms~\cite{bay2006surf, lowe1999, collet2011moped, hinterstoisser2011}. More recently, neural networks have been used to identify correspondences~\cite{xiang2017posecnn, kehl2017ssd, peng2019pvnet, park2019pix2pose}.
Current state-of-the-art methods often use the render-and-compare approach~\cite{cosypose, megapose, li2018deepim, zakharov2019dpod, nguyen2024gigapose} that learns to find the pose for which the rendered 3D mesh best matches the input image. Some methods apply multiview scene consistency refiner to improve single view results~\cite{cosypose}.
In contrast to these methods, we apply physical scene consistency (such as penetration or surface support), which requires pose estimates only from a single view. Our approach is agnostic to the choice of the specific pose estimation algorithm. 

\mypar{Differentiable collisions.}
Several approaches have been developed to address the challenges associated with computing the distance between objects and the associated distance derivatives with respect to object motion.
Escande et al.~\cite{escande2014strictly_convex} showed that the distance between objects is continuously differentiable for any pair of strictly convex objects and 
proposed an algorithm to generate strictly convex hulls and compute the collision derivative.
Werling et al.~\cite{werling2021fast} proposed a method to compute the collision derivative for arbitrary convex meshes, not necessarily strictly convex.
However, representing an object as a mesh makes the region around the vertices of the mesh locally flat, even when the original object has curved surfaces, which often leads to inaccurate gradients.
Montaut et al.~\cite{diffcol} addressed this problem by proposing DiffCol, which is an approach based on the Gilbert-Johnson-Keerthi algorithm~\cite{gilbert1988gjk} and randomized smoothing~\cite{berthet2020smoothing}.
This method compensates for the local approximation of the mesh representation, enhancing the accuracy and robustness of collision derivatives. We build on this work as a geometric foundation for our approach but focus on 6D object pose estimation for real-world scenes. 

\vspace{3mm}
\mypar{Collision resolution in human hand and body pose estimation.}
Modeling collisions is also important in the field of human pose estimation and hand pose estimation~\cite{rong2021handcoll,smith2020handelastic}.
The shape of the hand is often modeled as a deformable object which allows the collisions to be solved by bending the fingers and compressing the skin.
Similarly to the estimation of hand pose, the shape of the human body can also adapt to collisions, \eg, by using soft constraints via a collision penalty term in the pose estimation loss functions~\cite{jiang2020multihumanpose, hasson2021handobjectconsist, hassan2019scenehumanpose, zhang2020perceiving}.
Other methods directly prohibit collisions either by differential methods~\cite{davydov2024cloaf} or introduce a collision potential between body parts~\cite{belagiannis2014}.
Compared to these works, our method resolves the physical consistency of rigid objects, considering not only the collisions but also the gravity and use 6D pose estimates from images.

\mypar{Rigid object collision resolution for pose estimation.}
One way to deal with physical inconsistencies is to detect and discard problematic estimates~\cite{deng2022ppf-fix-ovelapping,fu2020}.
Instead of discarding colliding estimates, our method refines their poses to resolve the collision to improve object pose estimation accuracy.
Collision resolution can be framed as solving a nonlinear optimization problem that minimizes a non-penetration loss function. In order to obtain well-behaved derivatives for the colliding shapes, existing methods propose using various approximations of the scene objects, such as prescribed support functions~\cite{lee2023uncertain}, super quadrics~\cite{landgraf2021simstack}, and voxels~\cite{wada2020morefusion}. These methods typically also use depth maps as input to their algorithms. In our case, we directly work with a mesh-based representation of the objects building on the work of~\cite{diffcol} and only require RGB frames. 

\begin{figure*}
    \centering
    \includegraphics[width=\linewidth]{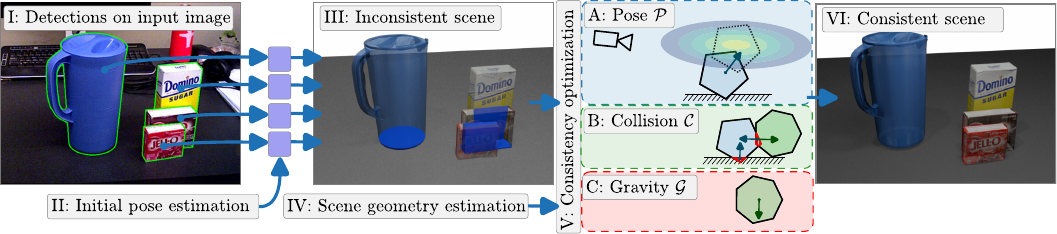}
    \caption{
    \textbf{The PhysPose pipeline} processes an input image by first (I)~detecting objects and then (II)~independently estimating their poses. This initial estimation often results in a (III)~physically inconsistent scene, where collisions are highlighted in blue. Our (V)~physically consistent optimization leverages (IV)~estimated scene geometry and  (A)~a~pose cost, (B)~a~collision cost, and (C)~a~gravity cost to generate a (VI)~physically consistent scene.
    (A)~Pose cost attracts the optimized object poses towards the initial measurements (indicated by dotted outlines). This cost accounts for the varying accuracy of depth estimation using non-isometric covariance, represented by the ellipsoid.
    (B)~Collision cost penalizes object penetration, visualized by red-shaded areas.
    (C)~Gravity cost minimizes the scene's potential energy.
    The gradients of these costs are represented by arrows and their formulas are provided in the appendix.
    }
    \label{fig:overview}
\end{figure*}

\section{Approach}
Given an image and a geometric description of the scene, our goal is to estimate the 6D pose of each object while enforcing physical consistency, ensuring that objects neither levitate nor collide with each other or the scene, as described first in Sec.~\ref{sec:epc}.
While scene geometry is often readily available in robotic applications (as it is required for other modules like motion planning), this is not the case in more general settings; therefore, we designed an approach to estimate a support table from two views, serving as an onboarding stage before pose estimation, as shown in Sec.~\ref{sec:table_pose}.

\subsection{Physically consistent object pose estimation}\label{sec:epc}
This section introduces a physically consistent object pose estimation framework that combines image data, scene geometry, and a novel optimization strategy. We begin by localizing objects within an input image using available methods like~\cite{nguyen2023cnos} and applying existing pose estimation methods, such as MegaPose~\cite{megapose} or FoundPose~\cite{ornek2024foundpose}, to each object crop independently. The resulting pose estimates, however, can exhibit inconsistencies by neglecting scene-level physical constraints. To address this, our core contribution is a novel optimization step, detailed in the remainder of this section and visualized in Fig.~\ref{fig:overview}, that enforces physical plausibility by penalizing object levitation and collisions while respecting the initial pose estimates and acknowledging the inherent difficulty of accurate depth estimation.

\newcommand{\percp}{\mathcal{P}}
\newcommand{\col}{\mathcal{C}}
\newcommand{\grav}{\mathcal{G}}

\mypar{Problem formulation.}
Our method takes as input: (i)~a~set of 3D meshes of movable objects, $\{O_i\}_{i=1}^N$; (ii)~a~set of 3D meshes representing static/environment objects, $\{S_i\}_{i=1}^M$; and (iii)~initial pose estimates for the movable objects obtained from an image-based pose estimator, denoted as $\meas{T}\sub{C}{Oi} \in \text{SE(3)}$ for object $O_i$ in the camera frame~$C$. We represent the pose as ${T}\sub{C}{Oi}$, comprising translation ${\vb*{t}}\sub{C}{Oi}$ and rotation ${R}\sub{C}{Oi}$. For the purpose of this section, the static object poses, ${T}\sub{C}{Si}$, are assumed to be known and fixed. Their estimation is discussed later. The goal is to refine the initial pose estimates, $\meas{T}\sub{C}{Oi}$, to produce physically plausible object arrangements.
This is achieved by minimizing a cost function that we define as a weighted sum of three components: a pose cost $\percp$, a collision cost $\col$, and a gravity cost $\grav$. These terms encourage pose proximity to initial estimates, prevent object overlap, and penalize levitation, respectively. Mathematically, the total cost, summed over all movable objects, is:
\begin{align}
    C = \sum_{i=1}^N \left( \percp_i + \zeta_\col \col_i + \zeta_\grav \grav_i \right), \label{eq:cost}
\end{align}
where $\zeta_\col$ and $\zeta_\grav$ are weights controlling the influence of the collision and gravity costs, respectively, and $N$ is the number of (movable) objects in the scene.
To minimize cost~\eqref{eq:cost}, we employ gradient-based optimization, using analytical gradients derived in the appendix. Next, we describe the individual cost terms in detail. 

\myparcontinuous{Pose cost~$\percp$} aims to keep the optimized object pose close to the image-based pose estimate while accounting for uncertainty. For the $i$-th object, we define the residual $\mathbf{e}_i$ between the optimized pose $T_{C,Oi}$ and the estimated pose $\tilde{T}_{C,Oi}$ as:
\begin{equation}
    \mathbf{e}_i = \left[\mathbf{t}_{C,Oi} - \mathbf{\tilde{t}}_{C,Oi}, \, \log(\tilde{R}_{C,Oi}^T R_{C,Oi})\right]^T \in \mathbb{R}^6,
\end{equation}
where $\mathbf{t}$ and $R$ represent the translation and rotation components of the pose, respectively, and $\log$ is the SO(3) logarithmic mapping~\cite{sola2021lie}. Using this residual, the pose cost for object $O_i$ is defined as the Mahalanobis distance between the optimized pose and the initial pose estimate:
\begin{equation} \label{eq:perc-cost}
    \mathcal{P}_i = \frac{1}{2} \norm{\mathbf{e}_i}_{\Sigma_{Ci}}^2 = \frac{1}{2}\mathbf{e}_i^T\Sigma_{Ci}^{-1}\mathbf{e}_i = \frac{1}{2}\mathbf{e}_i^TH_i\mathbf{e}_i,
\end{equation}
where $H_i = \Sigma_{Ci}^{-1}$ is the precision matrix, and $\Sigma_{Ci}$ is the covariance matrix defined in the camera frame.

\begin{figure}[t]
    \includegraphics[width=0.75\linewidth]{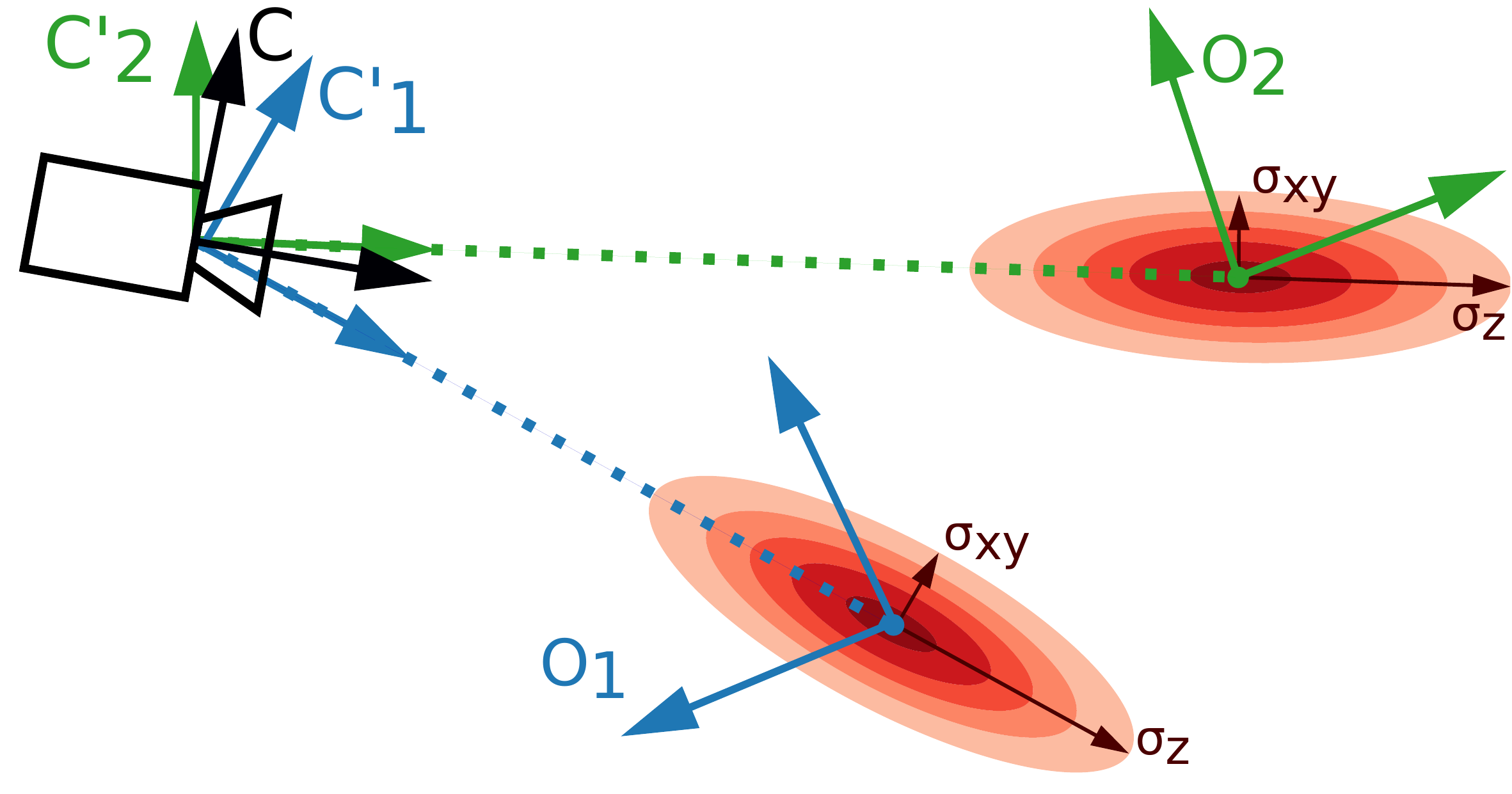}
    \caption{\textbf{Translation covariance ellipses.} Given two objects $O_1$ and $O_2$, we illustrate the covariance ellipses that we use in the pose cost. The standard deviation in the coordinate frames $C_{i}^{'}$ is larger along the axis pointing to the object, which reflects the difficulty of estimating depth from a single-view input image. This allows the objects to move more freely in this direction during the physics-based refinement optimization.}
    \label{fig:perc}
\end{figure}

To account for the inherent uncertainty in estimating the depth from the image input, we design a covariance matrix that allows for greater pose variation along the ray pointing toward the object.  Specifically, following the approach of~\cite{priban2024sampose}, we define a new camera coordinate frame $C_{i}'$ for each object, where the $z$-axis aligns with the vector from the camera to the object's center (see Fig.~\ref{fig:perc}).  In this coordinate frame, the translational covariance is a diagonal matrix:
\begin{equation}
    \Sigma^t_{C'i} = \text{diag}(\sigma_{xy}^2, \sigma_{xy}^2, \sigma_{z}^2),
\end{equation}
where $\sigma_{xy}$ and $\sigma_{z}$ represent the standard deviations in the $x, y,$ and $z$ directions, respectively.  These standard deviations are estimated from the real-world annotated data.  The rotational covariance is assumed to be diagonal in the object coordinate frame:
\begin{equation}
    \Sigma^R_{Oi} = \text{diag}(\sigma_{\theta}^2, \sigma_{\theta}^2, \sigma_{\theta}^2).
\end{equation}
Finally, the translational and rotational covariance matrices are transformed back to the original camera frame $C$ as:
\begin{align}
    \Sigma_{Ci}^t &= R_{C,C'i} \Sigma_{C'i}^t R_{C,C'i}^T \, ,  \\
    \Sigma_{Ci}^R &= R_{C,Oi} \Sigma_{Oi}^R R_{C,Oi}^T \, ,
\end{align}
where $R_{C,C'i}$ and $R_{C,Oi}$ are the rotation matrices from frame $C'$ to $C$ and from frame $O_i$ to $C$, respectively. The complete covariance matrix is then constructed as a block diagonal matrix:
$\Sigma_{Ci} = \text{diag}(\Sigma_{Ci}^t, \Sigma_{Ci}^R)$.

\myparcontinuous{Collision cost~$\col$} aims to resolve overlapping shapes by moving their poses into a non-colliding configuration. For the $i$-th movable object $O_i$, its collision cost $\mathcal{C}_i$ is computed as the sum of pairwise collision costs with both movable and static objects:
\begin{align}
    \mathcal{C}_i = \sum_{\substack{j=1 \\ j \neq i}}^{N} \mathcal{C}_{Oi,Oj} + \sum_{k=1}^{M} \mathcal{C}_{Oi,Sk},
\end{align}
where $N$ is the number of movable objects, $M$ is the number of static objects, $\mathcal{C}_{Oi,Oj}$ is the collision cost between movable objects $O_i$ and $O_j$, and $\mathcal{C}_{Oi,Sk}$ is the collision cost between movable object $O_i$ and static object $S_k$.

\begin{figure}[t]
    \includegraphics[width=\linewidth]{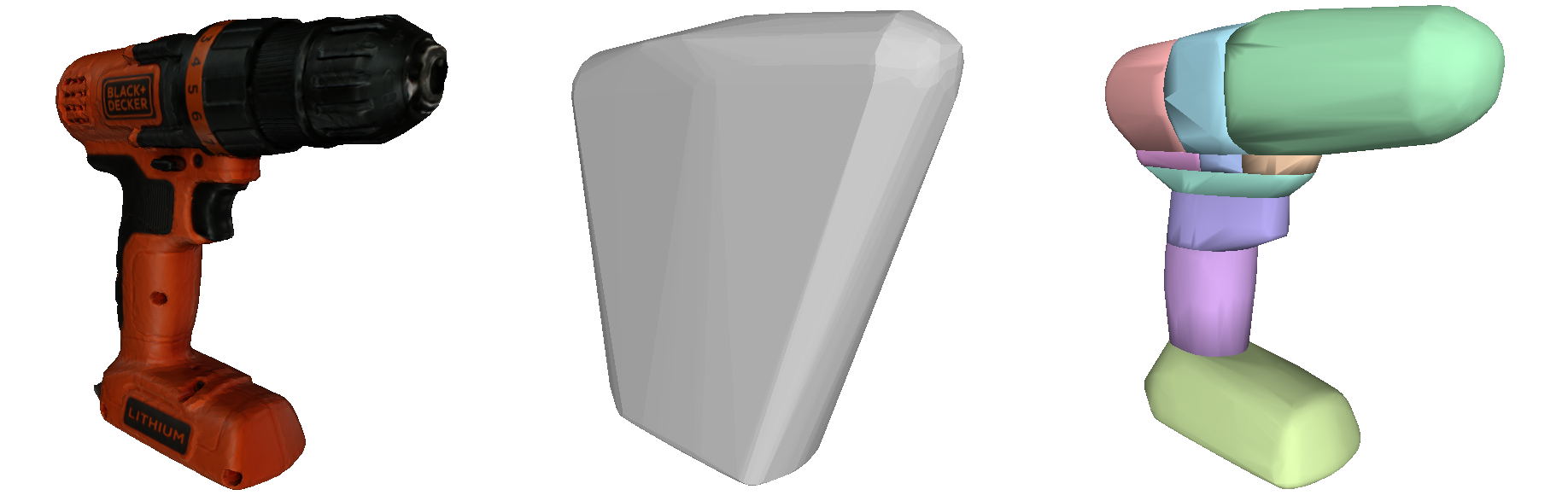}
    \caption{\textbf{Visualization of object convex approximations.} From left to right: the non-convex mesh from the YCB-V dataset, its convex hull, and a convex decomposition with individual parts distinguished by different colors.
    We use convex decomposition to obtain detailed geometry representation decomposed into convex subparts for which efficient differentiable collision distance computation is feasible.
    }
    \label{fig:decomp_hull}
\end{figure}

The pairwise collision cost between two objects, $\mathcal{A}$ and $\mathcal{B}$, depends on the signed distance between them at their current configurations. However, efficient and differentiable signed distance computation typically requires convex shapes. To preserve geometric detail, we apply an approximate convex decomposition method~\cite{wei2022approximate}, as illustrated in Fig.~\ref{fig:decomp_hull}. We then compute the signed distances between all pairs of subparts. 
We define $\mathcal{A}_i$ as the $i$-th convex part of object $\mathcal{A}$, where $\mathcal{A}_i \in \mathcal{A}$.
The pairwise collision cost is then computed as:
\begin{equation}
    \mathcal{C}_{\mathcal{A},\mathcal{B}} = \frac{1}{n_\text{col}}
    \sum_{\mathcal{A}_i \in \mathcal{A}} \sum_{\mathcal{B}_j \in \mathcal{B}} \left[ -d(\mathcal{A}_i, \mathcal{B}_j) \right]_+,
\end{equation}
where $d(\mathcal{A}_i, \mathcal{B}_j)$ is the signed distance between convex parts $\mathcal{A}_i$ and $\mathcal{B}_j$, $[x]_+ = \max(0, x)$ is the hinge loss, which penalizes only negative distances (collisions), and $n_\text{col}$ is the number of colliding pairs.

\myparcontinuous{Gravity cost~$\grav$} prevents objects from levitating by encouraging them to move towards static objects in the direction of gravity. We assume that the gravity direction is provided as part of the scene description. 
In practice, we infer the gravity direction from the scene's support table pose, assuming that the table surface is horizontal.

To compute the gravity cost for a movable object $\mathcal{A}$, we first identify the closest convex subpart $\mathcal{B}$ of a static object along the gravity direction using the signed distance function. The gravity cost is then computed as the average positive distance of each convex subpart of the movable object $\mathcal{A}$ to this closest static subpart $\mathcal{B}$:
\begin{align}
    \mathcal{G}_\mathcal{A} = \delta_\mathcal{A} \frac{1}{|\mathcal{A}|} \sum_{\mathcal{A}_i \in \mathcal{A}} [d(\mathcal{A}_i, \mathcal{B})]_+,   
\end{align}
where $\mathcal{A}_i$ represents the $i$-th convex subpart of object $\mathcal{A}$, $|\mathcal{A}|$ denotes the number of convex subparts in $\mathcal{A}$, $\mathcal{B}$ is the closest convex subpart of a static object, and $[x]_+ = \max(0, x)$ is the hinge loss used to penalize only positive distances.
To facilitate the optimization of scenes where objects rest on top of other movable objects, we introduce a term $\delta_\mathcal{A}$ that disables the gravity cost for a given object $\mathcal{A}$ when it is in collision with any other object in the scene.
This term indicates if an object has sufficient support from below to overcome gravity and remain stationary.

\begin{figure}[t]
    \centering
    \includegraphics[width=\linewidth]{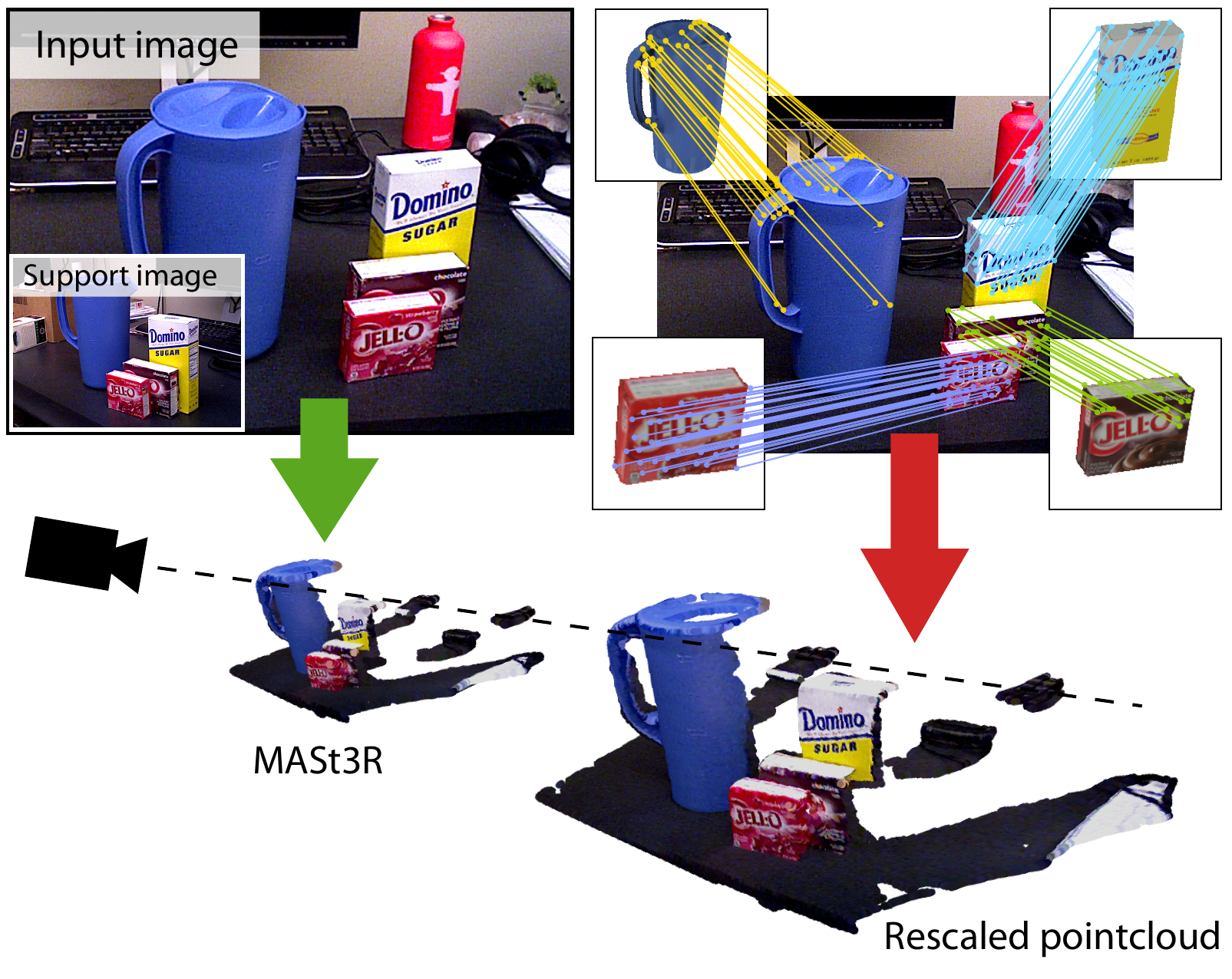}
    \caption{\textbf{Scaled point cloud computation for scene geometry estimation.}
    This work introduces a method to estimate metrically accurate scene geometry by leveraging the up-to-scale point cloud generated by MASt3R~\cite{mast3r_arxiv24}. Given an input image and support view (top-left), we utilize MASt3R to produce an initial reconstruction (indicated by green arrow). Then, we enforce metric calibration by establishing correspondences between the input image and rendered RGBD templates from the known 3D mesh of the object (top-right). This allows us to derive a single scale factor that transforms (red arrow) the MASt3R reconstruction into a metrically accurate point cloud for scene geometry estimation.    
    }
    \label{fig:master}
    \vspace{-3mm}
\end{figure}

\subsection{Scene geometry estimation}\label{sec:table_pose}
With the recent progress in 3D computer vision~\cite{mast3r_arxiv24,wang2024dust3r,colmap1,colmap2}, it is becoming possible to estimate the scene geometry to provide the 3D model of the environment where physical consistency can be optimized. However, two key challenges remain: (i) how to make the scene geometry estimation robust to noise and outliers; and (ii) how to metrically calibrate the output and align it with the object pose estimates coming from the pose estimator. We develop an automatic procedure to address these issues and demonstrate it in estimating the 3D geometry of tabletops that are common in the pose estimation BOP benchmark datasets. Details are given next.

Accurately estimating the 6D pose of a table from RGB images is challenging, as tables often lack well-defined key points and exhibit minimal texture variation. Unlike many objects with known 3D models, tables must be inferred from their geometric properties alone. A single-view depth estimation approach often suffers from inaccuracies, leading to inconsistencies in the reconstructed scene. To achieve the level of precision required for physically plausible object poses, we instead leverage a two-view approach. Given an input image and one support image of the scene, we reconstruct a point cloud using MASt3R~\cite{mast3r_arxiv24}, which provides a dense representation of the scene. 
Outliers are removed using prediction confidence. However, the reconstruction remains on an arbitrary scale. 

To obtain a metrically accurate reconstruction, we compute a depth map from the MASt3R prediction and rescale it by aligning it with the depth obtained using the initial object pose predictions: We first render an RGBD template from the known 3D mesh of the object for all predicted objects in their respective poses for the input image and remove depth values near object boundaries for more robust fitting.
Then, we run SuperGlue~\cite{sarlin2020superglue} to get 2D correspondences on the objects in the input image and the rendered template and find the corresponding 3D points. After obtaining the 3D correspondences, we use RANSAC to estimate the scene rescaling factor by sampling a pair of 3D correspondences on a single object and computing a distance ratio of the two 3D points from the rendered depth, where the distances are accurate, and the two 3D points from MASt3R. This approach is robust even if the initial pose estimates contain some errors.

Once the depth map is rescaled, we extract the table plane using a robust RANSAC-based fitting procedure. We reconstruct the scene point cloud from the depth map while removing low-confidence pixels and pixels corresponding to predicted objects. Furthermore, we run GroundingDINO~\cite{liu2024grounding} object detector to detect the table in the image and remove any pixels outside of the predicted bounding box. After the point cloud reconstruction, we use RANSAC to estimate the plane equation parameters representing the table plane. We illustrate the table estimation approach in Fig.~\ref{fig:master}.

\begin{table}[tb]
    \centering
    \small
    \begin{tabular}{crcccc} \toprule
         &  Method & MSPD & MSSD & VSD & AR \\ \midrule
        \multirow{4}{*}{\rotatebox[origin=c]{90}{YCB-Video}}
          & MegaPose~\cite{megapose} & {0.728} & {0.597} & {0.535} & {0.620} \\
          & Ours with~\cite{megapose} & \textbf{0.735} & \textbf{0.730} & \textbf{0.658} & \textbf{0.707} \\ \cmidrule{2-6}
          & FoundPose~\cite{ornek2024foundpose} & {0.797} & {0.670} & {0.603} & {0.690} \\
          & Ours with~\cite{ornek2024foundpose} & \textbf{0.804} & \textbf{0.802} & \textbf{0.718} & \textbf{0.775} \\
        \midrule
        \multirow{4}{*}{\rotatebox[origin=c]{90}{HOPE-Video}}
          & MegaPose~\cite{megapose} & {0.381} & {0.330} & {0.347} & {0.353} \\
          & Ours with~\cite{megapose} & {\textbf{0.391}} & {\textbf{0.380}} & {\textbf{0.415}} & {\textbf{0.395}} \\ \cmidrule{2-6}
          & FoundPose~\cite{ornek2024foundpose} & {0.302} & {0.266} & {0.292} & {0.286} \\
          & Ours with~\cite{ornek2024foundpose} & {\textbf{0.307}} & {\textbf{0.317}} & {\textbf{0.348}} & {\textbf{0.324}} \\
        \bottomrule
    \end{tabular}
     \caption{
    \textbf{Results in known environments.}
    We analyze the effect of incorporating physical consistency into pose estimation by examining average recall for individual metrics (MSPD, MSSD, and VSD) and the overall average recall (AR). These results are specific to a known environment, mirroring a typical robotics setup, where the environment is known a priori. Our proposed PhysPose method refines initial pose estimates provided by two baselines: MegaPose~\cite{megapose} and FoundPose~\cite{ornek2024foundpose}.
    }\label{tab:known_scenes}
\end{table}

\begin{figure*}[t]
    \centering
    \small
    \begin{overpic}[width=\linewidth,tics=5]{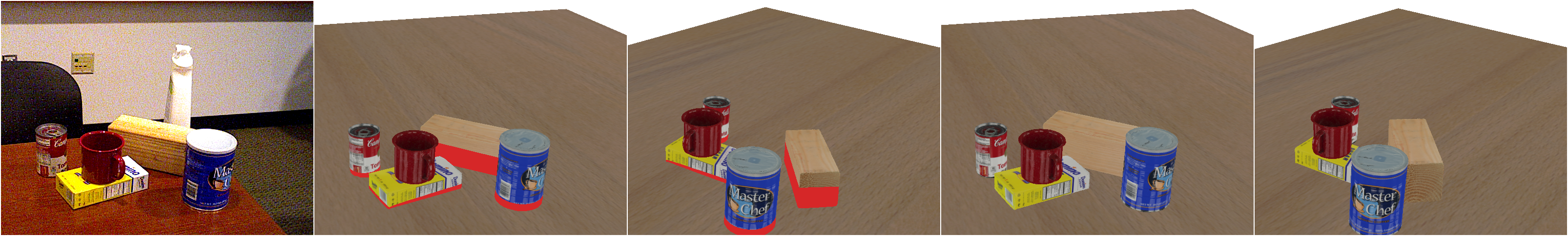}
        \put(0.1,13){\transparent{0.8}\colorbox{white}{\transparent{1}{Input image}}}
        \put(20.1,13){\transparent{0.8}\colorbox{white}{\transparent{1}{Initial~\cite{megapose} - camera view}}}        
        \put(40.1,13){\transparent{0.8}\colorbox{white}{\transparent{1}{Initial~\cite{megapose} - side view}}}
        \put(60.1,13){\transparent{0.8}\colorbox{white}{\transparent{1}{Ours - camera view}}}
        \put(80.1,13){\transparent{0.8}\colorbox{white}{\transparent{1}{Ours - side view}}}
    \end{overpic}
    \includegraphics[width=\linewidth]{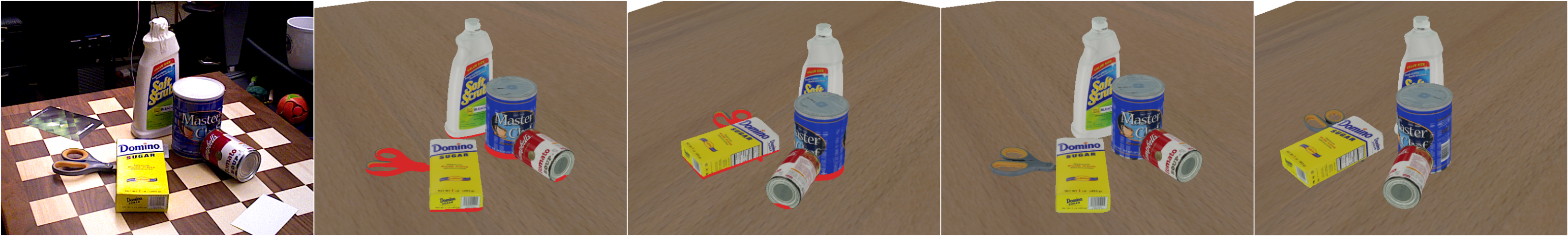}
    \includegraphics[width=\linewidth]{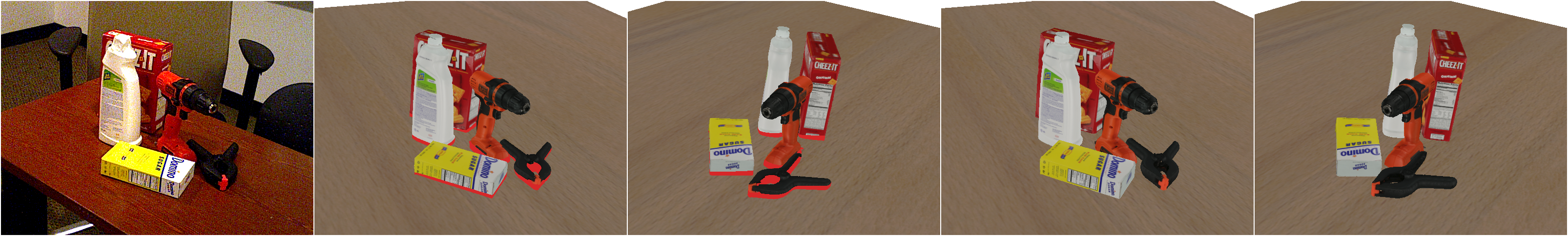}
    \includegraphics[width=\linewidth]{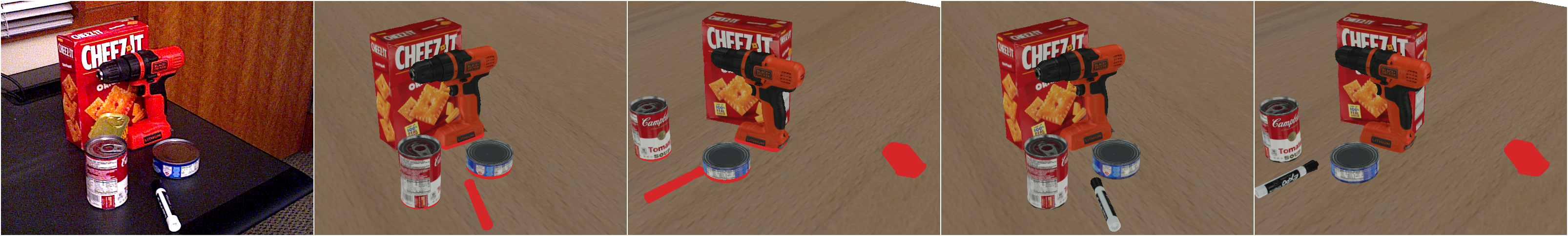}
    \caption{\textbf{Qualitative results on YCB-V datasets.}
    We estimate initial object poses from a single input image (1st column) using MegaPose~\cite{megapose}. The resulting scene, shown from camera (2nd column) and side (3rd column) views, exhibits significant physical inconsistencies, with colliding parts highlighted in red. Our physical consistency optimization method significantly reduces these collisions, leading to a more plausible scene arrangement shown from the camera view (4th column) and the side view (5th column). Notice how our method successfully resolves collisions in the first two rows even though the objects are placed on top of each other. {\bf Additional qualitative results are shown in the appendix.}
    }
    \label{fig:qualitative}
\end{figure*}

\section{Experiments}
In this section we introduce evaluation datasets and metrics, report results of our optimization-based pose refinement in known and estimated environments, ablate the importance of the different terms in our physics-based cost function, discuss the limitations of our method, and present an application of our approach to a real robot object grasping task. 

\noindent\textbf{Datasets.}
We quantitatively evaluate our method on two BOP benchmark~\cite{hodan2023bop} datasets: YCB-Video~\cite{xiang2017posecnn} and HOPE-Video~\cite{hopevideo}. YCB-Video is a real-world dataset containing Yale-CMU-Berkeley (YCB) objects, while HOPE-Video features household objects for pose estimation. Both are used for 6D pose estimation and capture tabletop scenarios.

\mypar{Metrics.}
Following the BOP Challenge evaluation protocol~\cite{hodan2023bop,hodavn2020bop}, we report average recall (AR) using three metrics: (i) Visible Surface Discrepancy (VSD)~\cite{hodavn2016evaluation}, which compares rendered distance maps, filtered by real-world depth data to consider only visible object parts; (ii) Maximum Symmetry-Aware Surface Distance (MSSD), measuring the maximum Euclidean distance between mesh vertices, accounting for object symmetry; and (iii) Maximum Symmetry-Aware Projection Distance (MSPD), similar to MSSD but measuring pixel distance after projecting vertices onto the image plane.

\begin{figure*}[t]
    \centering
    \small
    \begin{overpic}[width=0.2\linewidth,tics=5]{fig/anim/anim_50_701/000701}
        \put(0.1,67){\transparent{0.8}\colorbox{white}{\transparent{1}{Input image}}}
    \end{overpic}%
    \begin{overpic}[width=0.2\linewidth,tics=5]{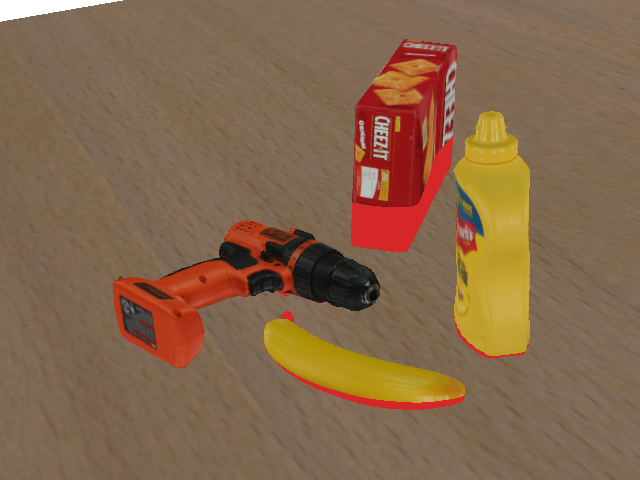}
        \put(0.1,67){\transparent{0.8}\colorbox{white}{\transparent{1}{Initial pose~\cite{megapose}}}}
    \end{overpic}%
    \includegraphics[width=0.2\linewidth]{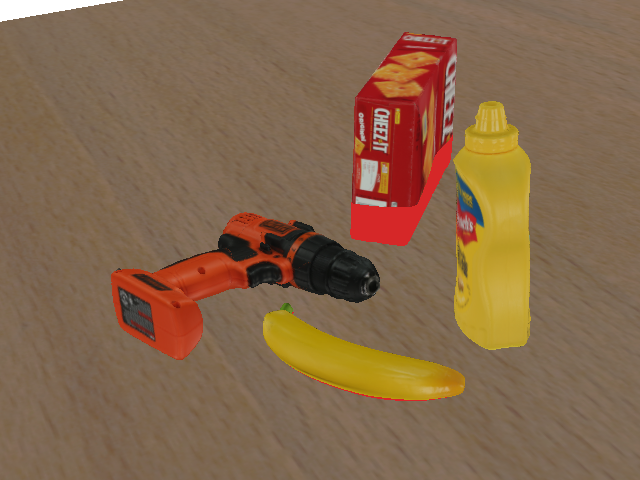}%
    \includegraphics[width=0.2\linewidth]{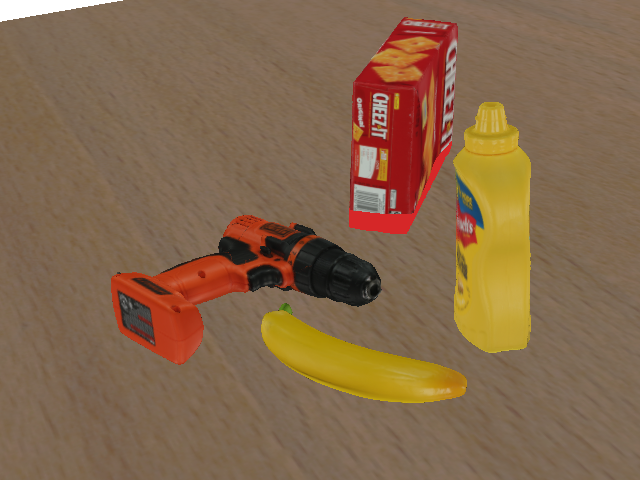}%
    \begin{overpic}[width=0.2\linewidth,tics=5]{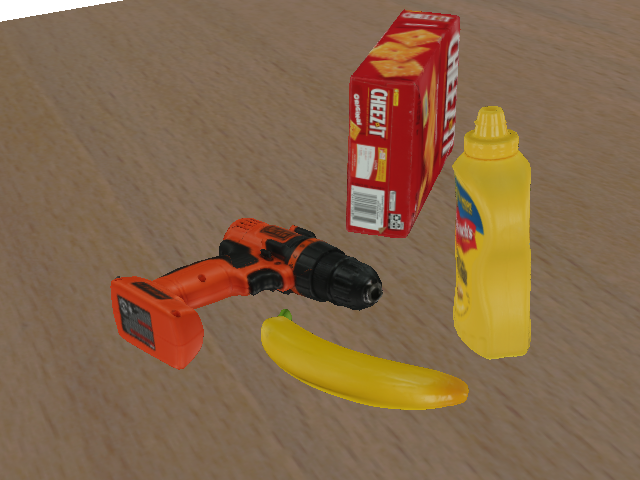}
        \put(0.1,67){\transparent{0.8}\colorbox{white}{\transparent{1}{Optimized pose}}}
    \end{overpic}\\
    \includegraphics[width=0.2\linewidth]{fig/anim/anim_57_1440/001440}%
    \includegraphics[width=0.2\linewidth]{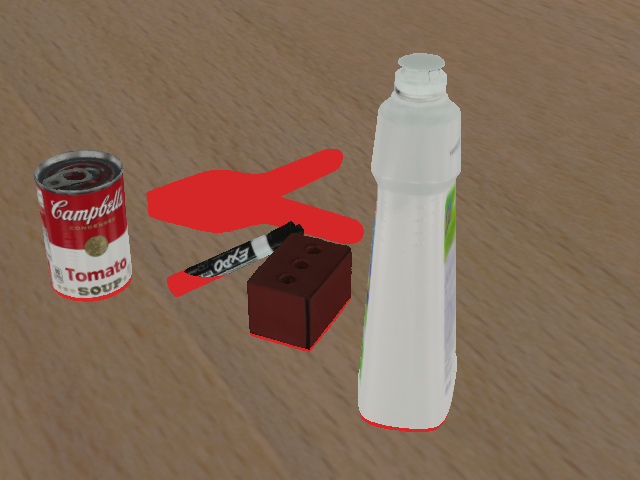}%
    \includegraphics[width=0.2\linewidth]{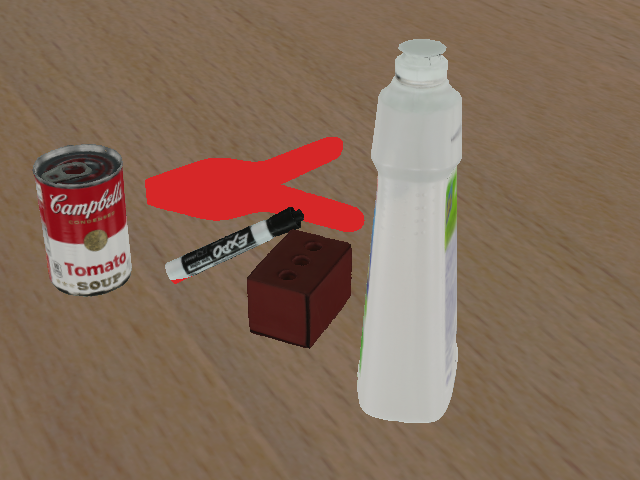}%
    \includegraphics[width=0.2\linewidth]{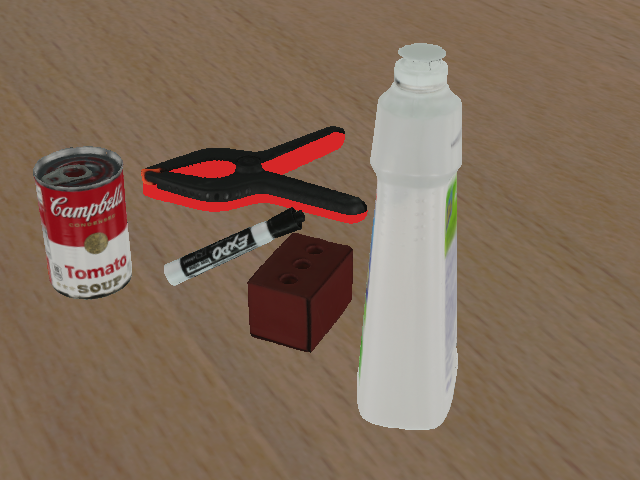}%
    \includegraphics[width=0.2\linewidth]{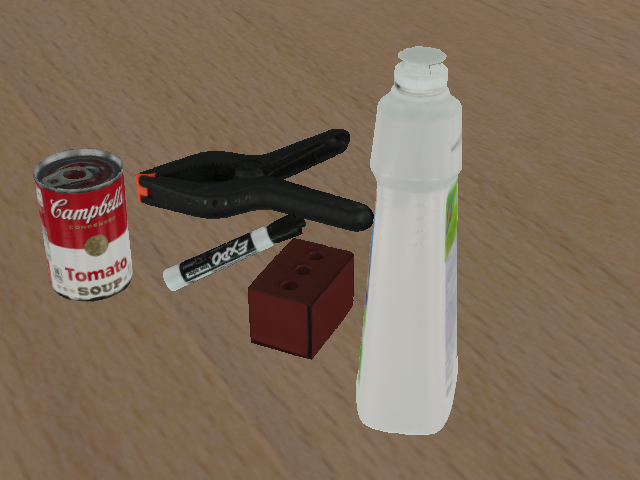}%
    \caption{
        \textbf{Physical consistency optimization.} Starting with an input image (left), initial object poses are estimated using MegaPose~\cite{megapose}. The optimization process then refines these poses from left to right, with intermediate results shown. Collisions are highlighted in red. Minor inconsistencies (\eg, banana in the first row, marker in the second row) are resolved early in the process, while larger collisions require more iterations. {\bf Videos illustrating the iterations of the optimization process are in the supplementary video~\cite{physpose_page}.} 
    }
    \label{fig:video}
    \vspace{-4mm}
\end{figure*}

\mypar{Results in known environments.}
Table~\ref{tab:known_scenes} presents results on the YCB-Video and HOPE-Video datasets for the case of known scene geometry, a common setup in robotics where scene models are often known and are used for tasks requiring, for example, collision checking. We compare our approach, which enforces physical consistency, with the MegaPose~\cite{megapose} and FoundPose~\cite{ornek2024foundpose} baselines, state-of-the-art methods designed for objects unseen during training. Enforcing physical consistency significantly improves the initial pose estimates. This is reflected in the BOP metrics; MSSD and VSD recall show notable gains, while MSPD recall remains relatively stable. This is because MSPD primarily measures projected error and is less sensitive to inaccuracies in depth estimation. The physical consistency constraints primarily aid in improving depth accuracy, explaining the gains observed in the 3D-sensitive MSSD and VSD metrics. We further validate our approach with qualitative results shown in Fig.~\ref{fig:qualitative} and Fig.~\ref{fig:video}.

\begin{table}[tb]
    \centering
    \small
    \setlength{\tabcolsep}{4pt}
    \begin{tabular}{lcccc} \toprule
          Scene estimation method &  MegaPose~\cite{megapose} & FoundPose~\cite{ornek2024foundpose} \\ \midrule
          a. Initial poses & 0.620 & 0.690\\
          b. Single view estimation  & 0.597 & 0.658\\
          c. With support view & \textbf{0.650} & \textbf{0.715} \\
          d. With known gravity & 0.648 & 0.710  \\  \midrule
         \rowcolor{lightgray} e.  Known scene & 0.706 & 0.774\\
          \bottomrule
    \end{tabular}
      \caption{ 
    \textbf{Results in the estimated environment.}
    Average recall on the YCB-Video dataset for the pose estimation using estimated scene geometry. The results compare the performance of MegaPose~\cite{megapose} and FoundPose~\cite{ornek2024foundpose} baselines with several scene geometry estimation strategies employed before physical consistency constraints are applied for pose refinement. The rows represent: (a.) Average recall using initial pose estimates; (b.) After incorporating scene geometry estimated from a single view only; (c.) After incorporating estimated scene geometry using an additional support view; (d.) After incorporating estimated scene geometry with a known gravity direction; and (e.) Average recall with known scene geometry (shown only for reference).
    }\label{tab:unknown_scenes}
\end{table}

\mypar{Results in estimated environments.}
We now evaluate the performance of our scene geometry estimation method (Sec.~\ref{sec:table_pose}) on the YCB-Video dataset, addressing situations where scene geometry is not known a priori. We present an ablation study (Table~\ref{tab:unknown_scenes}) that compares our full method (c.) to several variants: (a.) a baseline without enforcing physical consistency, (b.) a version of scene geometry estimation using only a single-view, and (d.) a version of scene geometry estimation that incorporates a known gravity direction (which in practice could be obtained from camera IMU data). In the single-view case (b.), we utilize Mast3r~\cite{mast3r_arxiv24} for monocular depth estimation, as a second supporting view may not always be available.
The results suggests that single-view geometry estimation is insufficient for improving pose accuracy, often degrading performance. In contrast, incorporating a support view for scene geometry estimation significantly boosted pose accuracy, exceeding initial pose estimates. Furthermore, leveraging a known gravity direction did not consistently enhance performance, suggesting it does not adequately regularize the estimated scene geometry in this context.

\mypar{Ablation of cost function terms.}
The effectiveness of our cost function is demonstrated through an ablation study, detailed in Table~\ref{tab:cost_abblation}. We report the average recall on the YCB-V dataset (in the known scene set-up) after selectively removing the pose~($\mathcal{P}$), collision~($\mathcal{C}$), and gravity~($\mathcal{G}$) cost terms. The results highlight the importance of each term in achieving better pose estimation accuracy via physical consistency. 

\begin{table}[tb]
    \centering
    \small
    \begin{tabular}{cccc} \toprule
        Pose cost~$\mathcal{P}$ & Collision cost~$\mathcal{C}$ & Gravity cost~$\mathcal{G}$ & AR \\ \midrule
       $\pmb{\checkmark}$ & $\pmb{\checkmark}$ & $\pmb{\checkmark}$ & \textbf{0.691} \\
        \texttimes & \checkmark & \checkmark & 0.540 \\
        \checkmark & \texttimes & \checkmark & 0.607 \\
        \checkmark & \checkmark & \texttimes & 0.615 \\
        \texttimes & \texttimes & \texttimes & 0.611 \\        
    \bottomrule
\end{tabular}
    \caption{
    \textbf{Ablation of different cost function terms.}
    We evaluate the impact of individual cost function terms on 10\% of the YCB-Video dataset, initialized with pose estimates from MegaPose~\cite{megapose}. The first row of the table presents our results with all cost terms active, while the last row shows the baseline MegaPose estimates (\ie, no active cost terms). Intermediate rows demonstrate the influence of disabling each cost term individually.
    \vspace{-5mm}
    }\label{tab:cost_abblation}
\end{table}

\begin{figure*}[t!]
    \centering
    \small
    \begin{overpic}[width=\linewidth,tics=5]{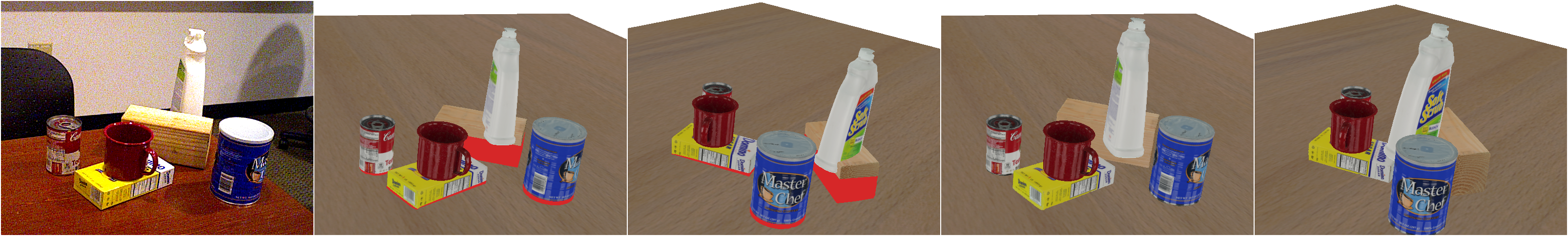}
        \put(0.1,13){\transparent{0.8}\colorbox{white}{\transparent{1}{Input image}}}
        \put(20.1,13){\transparent{0.8}\colorbox{white}{\transparent{1}{Initial~\cite{megapose} - camera view}}}        
        \put(40.1,13){\transparent{0.8}\colorbox{white}{\transparent{1}{Initial~\cite{megapose} - side view}}}
        \put(60.1,13){\transparent{0.8}\colorbox{white}{\transparent{1}{Ours - camera view}}}
        \put(80.1,13){\transparent{0.8}\colorbox{white}{\transparent{1}{Ours - side view}}}
    \end{overpic}
    \includegraphics[width=\linewidth]{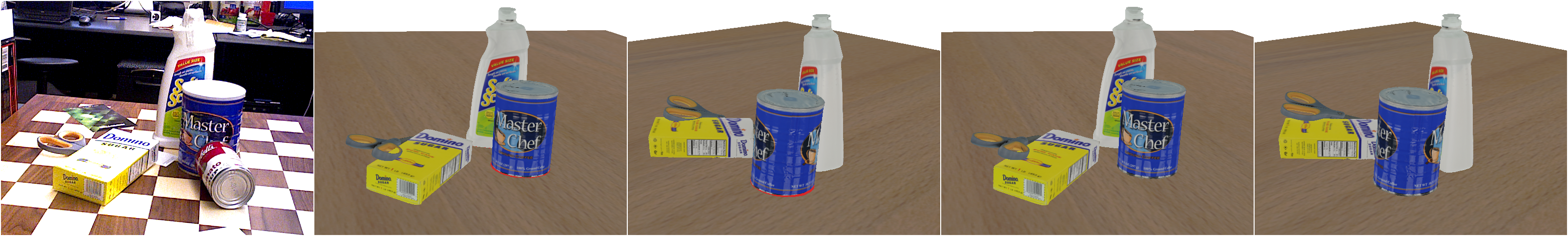}
    \caption{
    \textbf{Failure cases.}
    Our method can fail when initial pose estimates result in significant object collisions. For example, large initial collisions can lead to incorrect object ordering along the optical axis during optimization, increasing the error. This behavior was observed in instances such as the wooden cube and bottle collision (top row), and also with the scissors and sugar box (bottom row).
    }
    \label{fig:bop_mp_vs_coll_fail}
\end{figure*}
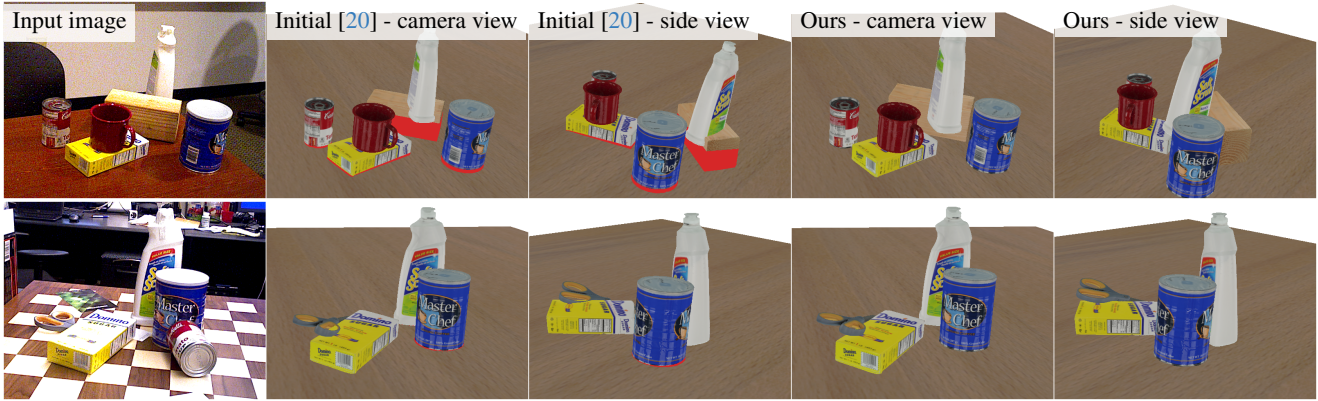

\mypar{Limitations.}
While our refinement significantly improves pose estimation, several failure modes remain. The simultaneous optimization of all objects makes the approach sensitive to incorrect object detections. False positives, for instance, can displace other objects, while false negatives, coupled with simulated gravity, may cause unsupported objects to fall. This contrasts with methods that treat each object independently. Incorrectly estimated poses for correctly detected objects can similarly propagate errors. These typical failures are illustrated in Fig.~\ref{fig:bop_mp_vs_coll_fail}.

\mypar{Application to robotics grasping.}
We conducted a pick-and-place robotics experiment using a  Panda robot to demonstrate the practical benefits of our pose refinement method. The task involves grasping an object and moving it to a designated location.
To create a challenging pose estimation scenario, we manually defined grasp handles for each object in the dataset, positioning them approximately 5~mm inside the object's surface. We captured top-view RGB images of the objects using a camera mounted on the robot gripper. This top-down perspective, common in industrial applications, presents a significant challenge for single-view pose estimation.
Object detection was done with Mask-RCNN~\cite{he2017maskrcnn}, and initial pose was estimated with~\cite{megapose}. Our approach was then applied on the initial pose estimates to incorporate physical consistency.

\begin{figure}[htb]
    \centering
    \begin{overpic}[width=\linewidth,tics=5]{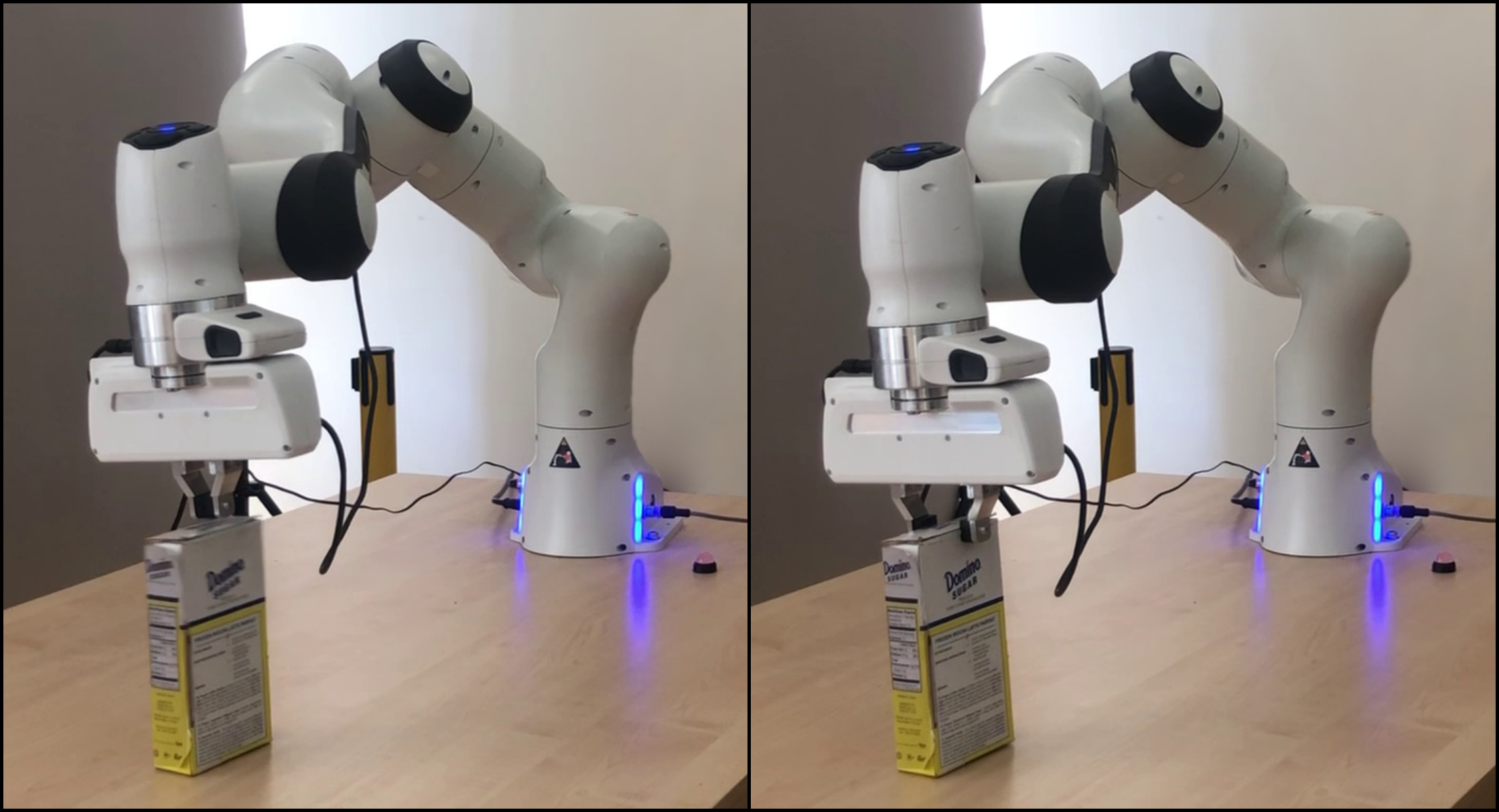}
        \put(0,50){\transparent{0.8}\colorbox{white}{\transparent{1}{Initial pose~\cite{megapose}}}}
        \put(50,50){\transparent{0.8}\colorbox{white}{\transparent{1}{Ours}}}
    \end{overpic}

    \caption{\textbf{Robotic grasping experiment.} The left image shows an unsuccessful grasp attempt of an object, with the pose estimated with~\cite{megapose}.
    Because initial pose is inaccurate, the robot closed the gripper a few millimeters above the object.
    Our pose refinement method provides a more accurate object pose estimate enabling the successful grasp shown on the right. {\bf Additional results are in the supplementary video~\cite{physpose_page}.}
    \vspace{-3.3mm}
    }
    \label{fig:robotic-experiment-grasp}
\end{figure}

We attempted to grasp each object using both the initial poses~\cite{megapose} and the refined poses generated by our approach, targeting the pre-defined handles. A collision-free grasping trajectory was planned using RRT~\cite{LaValle2000rrt}, and the motion was executed using position control.
We repeated the experiment five times for three distinct objects from the YCB-Video dataset (cracker box, mustard bottle, and sugar box), measuring the grasp success rate. Grasping attempts based on initial pose estimates completely failed for the cracker and sugar boxes and only achieved a 60\% success rate with the mustard bottle. In contrast, our method enabled successful grasping of all three objects with an 80\% success rate. This is thanks to the significantly improved pose estimates that enable more accurate handling of the objects. Fig.~\ref{fig:robotic-experiment-grasp} illustrates examples of successful and unsuccessful grasps. 

\section{Conclusion}
We present an optimization procedure that enforces physical consistency in a scene where object poses are initially estimated using any model-based pose estimation method. The minimized loss function consists of three terms: an initial pose cost keeps the objects close to their initial pose estimates while accounting for their inherent distance uncertainty; a collision cost resolves collisions between objects and the environment; and a gravity cost resolves levitating objects by modeling gravity.
We also develop an RGB-only metric plane estimation procedure using a two-view scene reconstruction algorithm coupled with a robust metric scale estimation.
Our method is general and can be used in conjunction with any pose estimator.
Quantitative experiments on YCBV and Hope-Video BOP datasets demonstrate that our method is able to improve the quality of recent RGB pose estimation methods, showing the importance of physical consistency for object pose estimation.
In addition, we conducted an experiment on the Franka Emika Panda robot for a pick-and-place task in a challenging setup.
Using our physically consistent object poses results in a significant increase in robustness for object grasping.
This work opens up the possibility of detailed physical reasoning about object configurations in real-world scenes.


\subsection*{Acknowledgments}
This work was partly supported by the Ministry of Education,
Youth and Sports of the Czech Republic through the e-INFRA CZ
(ID:90254), and by the European Union’s Horizon Europe projects AGIMUS (No. 101070165), euROBIN (No. 101070596), and ERC FRONTIER (No. 101097822).

Views and opinions expressed are however those of the author(s) only and do not necessarily reflect those of the European Union or the European Research Council. Neither the European Union nor the granting authority can be held responsible for them.

{
    \small
    \bibliographystyle{ieeenat_fullname}
    \bibliography{ref}
}

\section*{Appendix}
\appendix


\section{Computation of analytic gradients}
The proposed approach applies gradient descent to minimize the total cost defined in the main paper in Eq.~(1).
To achieve that, we derive analytical gradients for each partial cost defined in Sec~3.1.


\paragraph{The pose gradient} for $i$-th object, denoted as $\grad \percp_i\itr{}$, guides the optimization process to maintain the object's pose close to the image-based estimate. It is computed as:
\begin{align}
    \grad \percp_i\itr{} &= \vb*{e}_i\itr{}^T H_i J_i\itr{}, 
\end{align}
where $\vb*{e}_i$ represents the residual vector between the optimized pose and the initial pose, defined as $\vb*{e}_i = [\vb*{t}_{C,Oi} - \vb*{\tilde{t}}_{C,Oi}, \, \log(\tilde{R}_{C,Oi}^T R_{C,Oi})]^T \in \mathbb{R}^6$, where $\vb*{t}_{C,Oi}$ and $\vb*{\tilde{t}}_{C,Oi}$ are the translation components of the optimized and estimated poses respectively, and $\log(\tilde{R}_{C,Oi}^T R_{C,Oi})$ is the logarithmic mapping of the rotation difference.  The term $H_i$ is the precision matrix, which is the inverse of the covariance matrix, $\Sigma\sub{Ci}$.  Finally, $J_i$ is the Jacobian matrix, given by $J_i\itr{} = \begin{bmatrix} R\sub{C}{Oi}\itr{} & \vb{0} \\ \vb{0} & \pdv{\log(R\sub{\meas{O}i}{Oi}\itr{})}{R\sub{\meas{O}i}{Oi}} \end{bmatrix}$, where $R\sub{C}{Oi}$ is the rotation matrix from the object $O_i$ to the camera frame $C$, and the bottom right component is the partial derivative of the logarithmic rotation difference with respect to the rotation. An analytical formula for the jacobian of the SO(3) $\log$ map can be found in~\cite{sola2021lie}, Appendix B,C, Eq.~(144), notated as $J_r(\theta)^{-1}$.

\paragraph{Collision gradient} between two objects, denoted as $\col_{\mathcal{A},\mathcal{B}}$, aims to resolve overlapping shapes by moving them into a non-colliding state. 
To obtain the collision gradient $\grad \col_{\mathcal{A},\mathcal{B}}$, we need to differentiate the pairwise collision cost from Section~3.1 with respect to the pose of the object $\mathcal{A}$ (denoted as $T_\mathcal{A}$). The derivative is:
\begin{align}
    \grad \col_{\mathcal{A},\mathcal{B}} = \frac{\partial \col_{\mathcal{A},\mathcal{B}}}{\partial T_{\mathcal{A}}} =  \frac{1}{n_\text{col}} \sum_{\mathcal{A}_i \in \mathcal{A}} \sum_{\mathcal{B}_j \in \mathcal{B}} \frac{\partial}{\partial T_{\mathcal{A}}} \left[ -d(\mathcal{A}_i, \mathcal{B}_j) \right]_+ \, .
\end{align}
This can be further expressed as:
\begin{align}
    \grad \col_{\mathcal{A},\mathcal{B}} = \frac{1}{n_\text{col}} \sum_{\mathcal{A}_i \in \mathcal{A}} \sum_{\mathcal{B}_j \in \mathcal{B}}
    \begin{cases}
        0 & \text{if } d(\mathcal{A}_i, \mathcal{B}_j) \geq 0 \\
        -\frac{\partial d(\mathcal{A}_i, \mathcal{B}_j)}{\partial T_{\mathcal{A}}} & \text{if } d(\mathcal{A}_i, \mathcal{B}_j) < 0
    \end{cases}
\end{align}
In our approach, the derivative of the signed distance, $\frac{\partial d(\mathcal{A}_i, \mathcal{B}_j)}{\partial T_{\mathcal{A}}}$, is obtained by the randomized smoothing approach described in~\cite{diffcol}.

\paragraph{Gravity gradient} for object $\mathcal{A}$, denoted as $\grav_\mathcal{A}$, prevents objects from levitating by encouraging them to move towards static objects in the direction of gravity.  It is computed based on the closest convex subpart $\mathcal{B}$ of a static object and the average positive distance of the movable object $\mathcal{A}$'s convex subparts to $\mathcal{B}$ as defined in Section~3.1.
The gravity gradient is then:
\begin{align}
    \grad \grav_\mathcal{A} = \frac{\partial \grav_\mathcal{A}}{\partial T_\mathcal{A}} = \frac{\partial}{\partial T_\mathcal{A}} \left( \delta_\mathcal{A} \frac{1}{|\mathcal{A}|} \sum_{\mathcal{A}_i \in \mathcal{A}} [d(\mathcal{A}_i, \mathcal{B})]_+ \right) \, .
\end{align}
Since $\delta_\mathcal{A}$ is a binary variable depending on the collision state and is assumed not to be directly dependent on the pose of object A (it depends on the state of other objects), and $|\mathcal{A}|$ is constant, we can rewrite the derivative as:
\begin{align}
    \grad \grav_\mathcal{A} = \frac{\delta_\mathcal{A}}{|\mathcal{A}|} \sum_{\mathcal{A}_i \in \mathcal{A}} \frac{\partial}{\partial T_\mathcal{A}} [d(\mathcal{A}_i, \mathcal{B})]_+ \, ,
\end{align}
where the partial derivative of the hinge loss is computed in the same way as for the collision gradient described above.

\section{Additional qualitative results}
\begin{figure*}[t]
    \centering
    \small
    \begin{overpic}[width=\linewidth,tics=5]{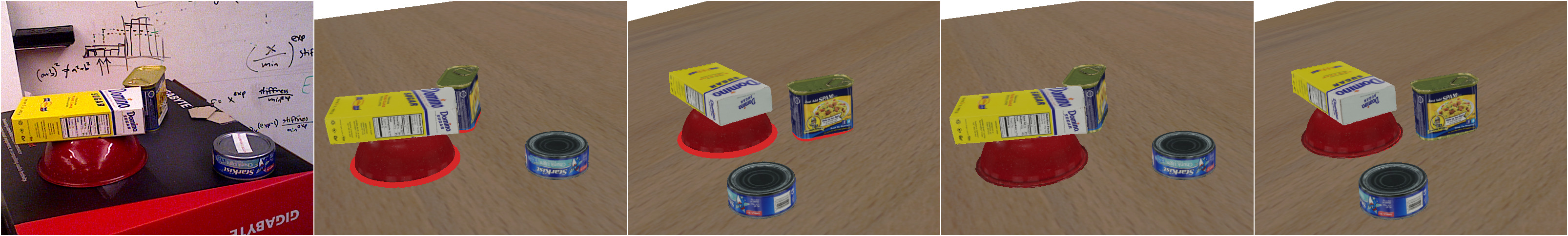}
        \put(0.1,13){\transparent{0.8}\colorbox{white}{\transparent{1}{Input image}}}
        \put(20.1,13){\transparent{0.8}\colorbox{white}{\transparent{1}{Initial~\cite{megapose} - camera view}}}        
        \put(40.1,13){\transparent{0.8}\colorbox{white}{\transparent{1}{Initial~\cite{megapose} - side view}}}
        \put(60.1,13){\transparent{0.8}\colorbox{white}{\transparent{1}{Ours - camera view}}}
        \put(80.1,13){\transparent{0.8}\colorbox{white}{\transparent{1}{Ours - side view}}}
    \end{overpic}
\includegraphics[width=\linewidth]{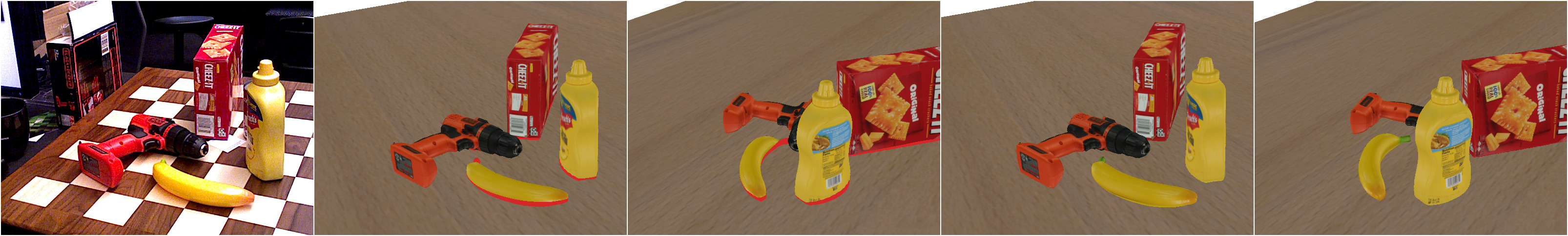}
\includegraphics[width=\linewidth]{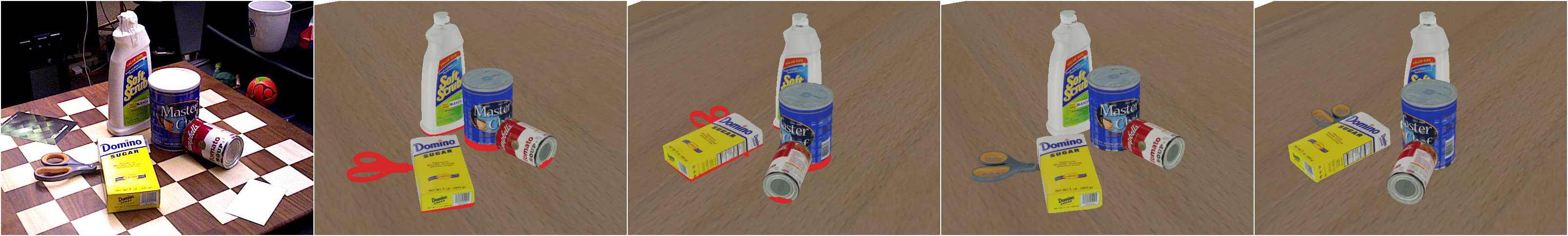}
\includegraphics[width=\linewidth]{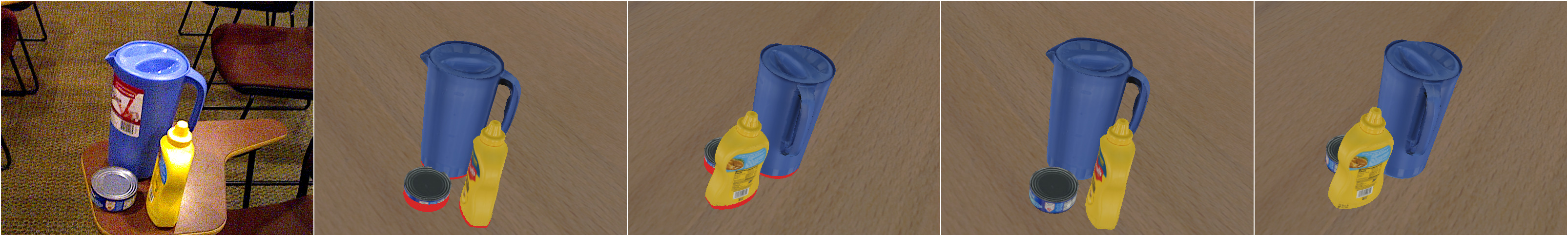}
\includegraphics[width=\linewidth]{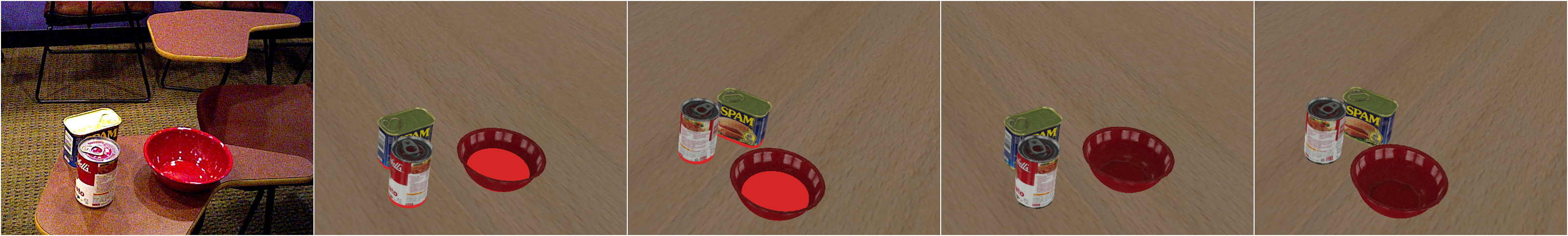}
\includegraphics[width=\linewidth]{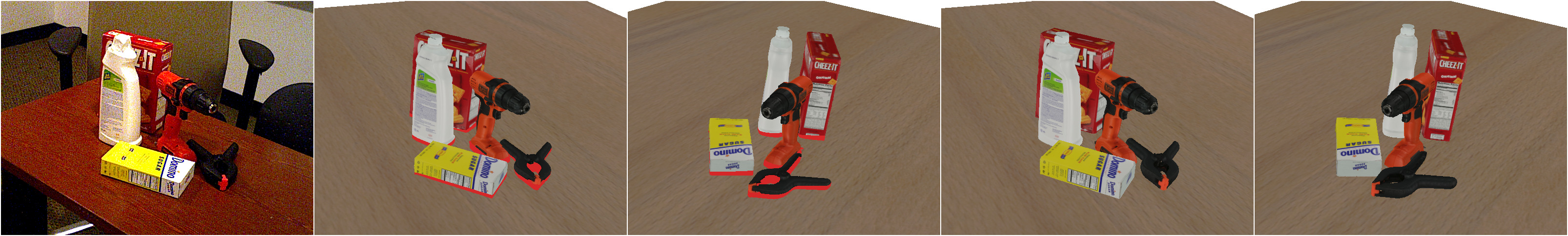}
\includegraphics[width=\linewidth]{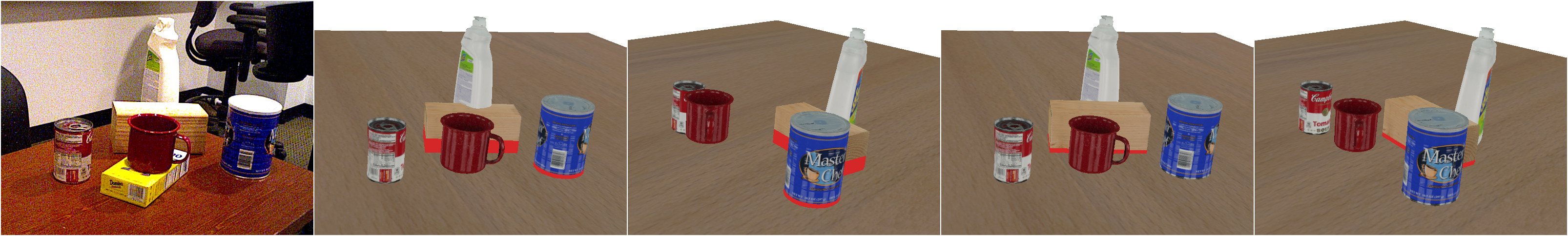}
   \caption{\textbf{Qualitative results on the YCB-V dataset.}
    We estimate initial object poses from an input image using MegaPose~\cite{megapose}. The resulting scene, shown from camera and side views, exhibits significant physical inconsistencies, with colliding parts highlighted in red. Our physical consistency optimization method significantly reduces these collisions, leading to a more plausible scene arrangement. Notice how our method successfully resolves collisions in the first two rows even though the objects are placed on top of each other. 
    }
    \label{fig:qualitative1}
\end{figure*}

\begin{figure*}[t]
    \centering
    \small
    \begin{overpic}[width=\linewidth,tics=5]{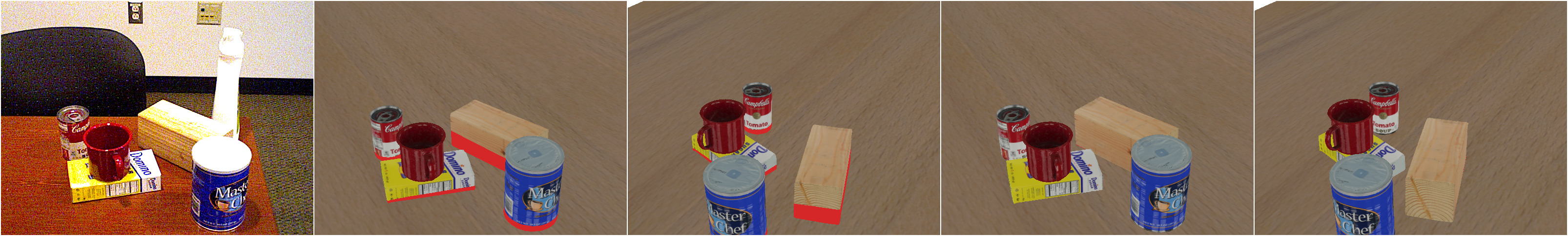}
        \put(0.1,13){\transparent{0.8}\colorbox{white}{\transparent{1}{Input image}}}
        \put(20.1,13){\transparent{0.8}\colorbox{white}{\transparent{1}{Initial~\cite{megapose} - camera view}}}        
        \put(40.1,13){\transparent{0.8}\colorbox{white}{\transparent{1}{Initial~\cite{megapose} - side view}}}
        \put(60.1,13){\transparent{0.8}\colorbox{white}{\transparent{1}{Ours - camera view}}}
        \put(80.1,13){\transparent{0.8}\colorbox{white}{\transparent{1}{Ours - side view}}}
    \end{overpic}
\includegraphics[width=\linewidth]{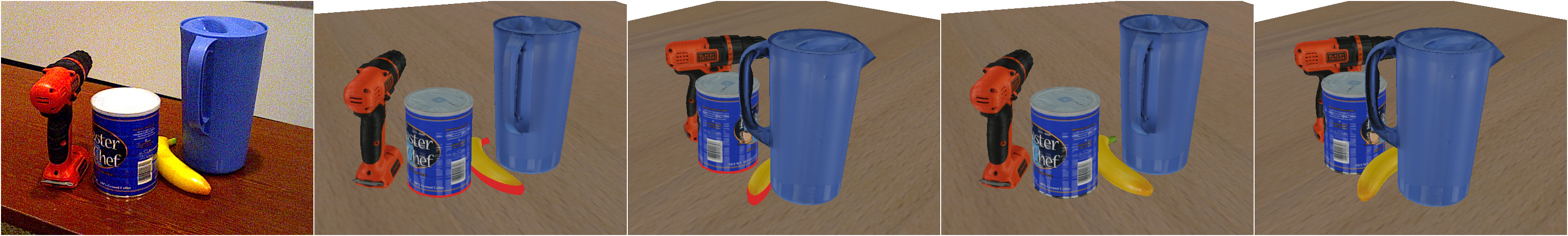}
\includegraphics[width=\linewidth]{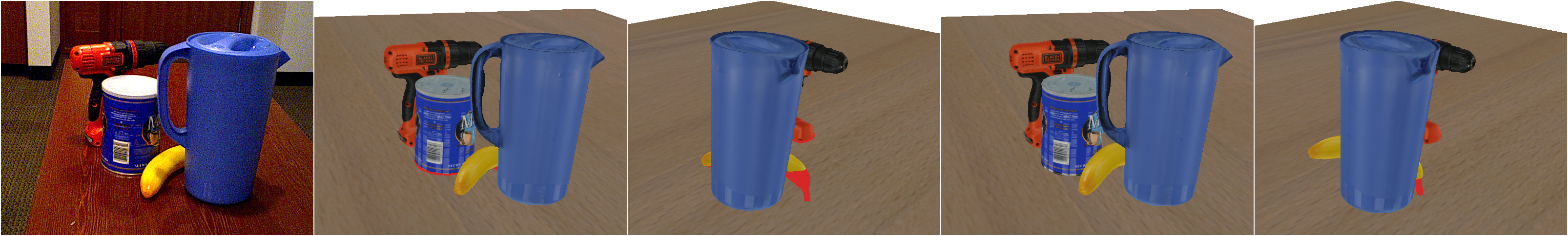}
\includegraphics[width=\linewidth]{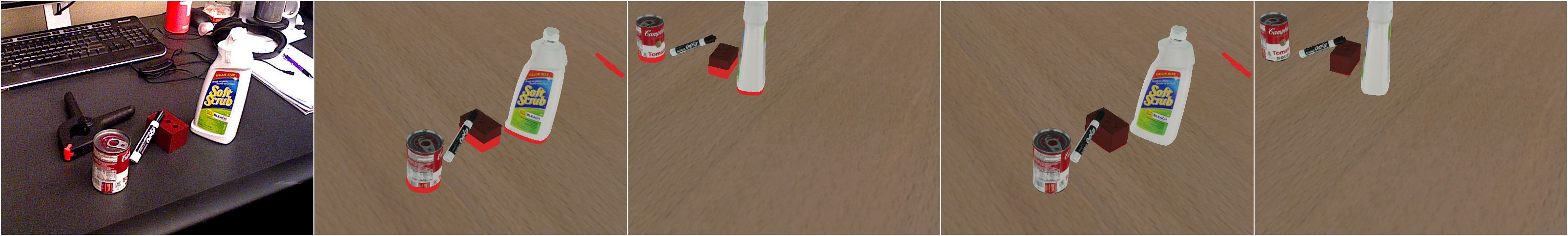}
\includegraphics[width=\linewidth]{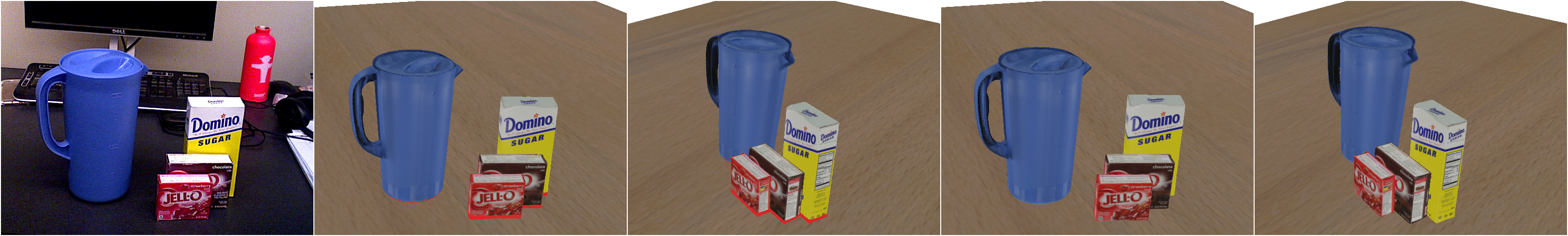}
\includegraphics[width=\linewidth]{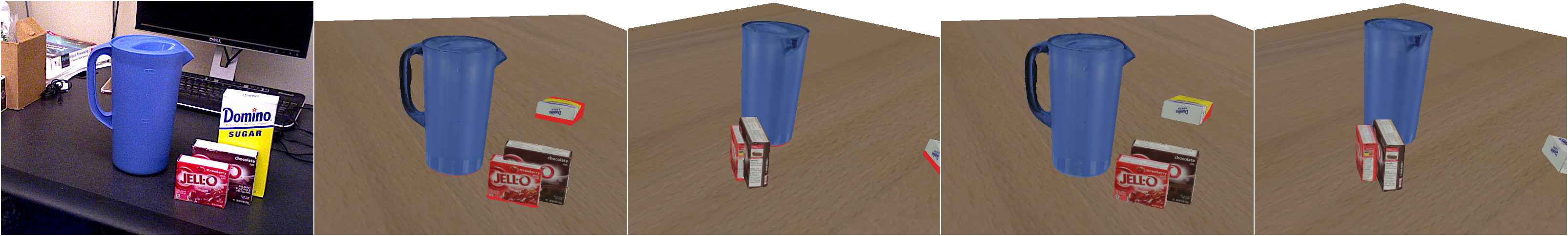}
\includegraphics[width=\linewidth]{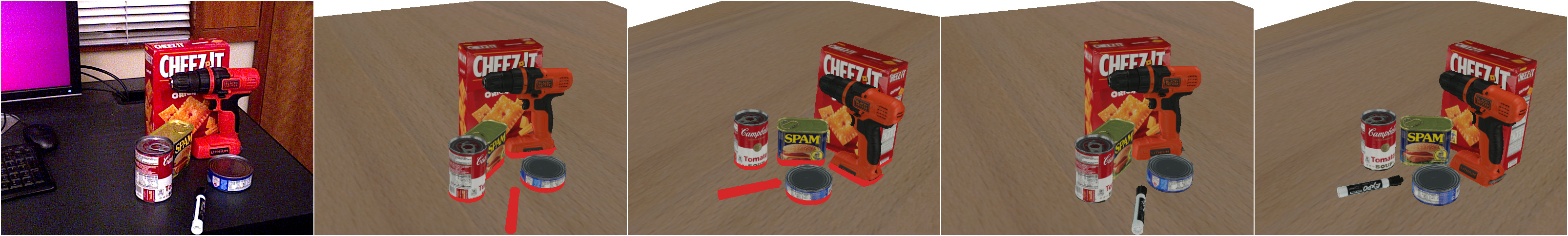}
    \caption{\textbf{Qualitative results on the YCB-V dataset.}
    We estimate initial object poses from an input image using MegaPose~\cite{megapose}. The resulting scene, shown from camera and side views, exhibits significant physical inconsistencies, with colliding parts highlighted in red. Our physical consistency optimization method significantly reduces these collisions, leading to a more plausible scene arrangement. Notice how our method successfully resolves collisions in the first two rows even though the objects are placed on top of each other. 
    }
    \label{fig:qualitative2}
\end{figure*}

In Fig.~\ref{fig:qualitative1} and Fig.~\ref{fig:qualitative2} we present additional qualitative results for enforcing physical pose estimation on the YCB-Video dataset.
Please note, that the HOPE-Video dataset features levitating objects in its initial poses, a characteristic not readily apparent in static visualizations. Therefore, the static visualization of qualitative results for the HOPE-Video dataset has been omitted. Please refer to the supplementary video, described below, for a dynamic visualization of the optimization process on a HOPE-Video scene.

\section{Supplementary video}
The first part of the supplementary video demonstrates the optimization process for scenes from the YCB-Video and HOPE-Video datasets. YCB-Video scenes, which initially exhibit object-object and object-scene collisions, are successfully resolved by our PhysPose method. 
For the HOPE-Video datasets, the initial poses exhibit levitation of objects above the tabletop; our method successfully mitigates this issue by attracting the objects downwards towards the scene geometry.

The second section of the video presents our robotic grasping experiments, contrasting the performance of the baseline method~\cite{megapose} with our approach. In the first experiment, the baseline's insufficient pose accuracy prevents a firm grasp of objects, as exemplified by the mustard bottle. Conversely, our method, leveraging its refined pose estimates, achieves successful grasps. Subsequent experiments demonstrate the baseline attempting grasps of objects predicted to collide with the scene geometry, potentially damaging the detected Cheez-It box. Note the Cheez-It box is quickly removed by the robot operator just before it would be damaged by the robot. Our method, however, effectively avoids these collisions and grasps the Cheez-It box without incident. Finally, the baseline occasionally predicts objects as levitating above the surface, leading to grasping attempts that miss the object entirely. This issue is rectified by our more accurate pose estimates, as evidenced by our successful grasp of the sugar box.

\end{document}